%% file: iclr2013_JADL.tex
\documentclass{article} 
\usepackage{nips12submit_e,times}
\usepackage{amsmath,amssymb}
\usepackage{algorithmic,algorithm}
\usepackage[dvips]{graphicx}
\usepackage{subfigure}
\usepackage{authordate1-4}
\usepackage{floatflt}

\def\D{{\mathbf D}}

\def\x{{\mathbf x}}

\def\a{{\mathbf a}}

\def\d{{\mathbf d}}

\def\R{{\mathbb R}}
\def\N{{\mathbb N}}

\def\Id{{\mathbf I}}

\newcommand{\argmin}{\operatornamewithlimits{argmin}}

\title{Jitter-Adaptive Dictionary Learning - Application to Multi-Trial Neuroelectric Signals}

\author{
Sebastian Hitziger\\
Project-Team Athena\\
INRIA Sophia Antipolis, France\\
\texttt{sebastian.hitziger@inria.fr} \\
\And
Maureen Clerc \\
Project-Team Athena\\
INRIA Sophia Antipolis, France\\
\texttt{maureen.clerc@inria.fr} \\
\And
Alexandre Gramfort\\
Institut Mines-Telecom, Telecom ParisTech, CNRS LTCI\\
\texttt{alexandre.gramfort@telecom-paristech.fr} \\
\And
Sandrine Saillet\\
Institut de Neurosciences des Syst\`{e}mes\\
UMR 1106 INSERM\\
Aix-Marseille Universit\'{e}\\
Facult\'{e} de M\'{e}decine La Timone\\
Marseille, France\\
\texttt{ssaillet53@gmail.com} \\
\And
Christian B\'{e}nar\\
Institut de Neurosciences des Syst\`{e}mes\\
UMR 1106 INSERM\\
Aix-Marseille Universit\'{e}\\
Facult\'{e} de M\'{e}decine La Timone\\
Marseille, France\\
\texttt{christian.benar@univmed.fr} \\
\And
Th\'{e}odore Papadopoulo \\
Project-Team Athena\\
INRIA Sophia Antipolis, France\\
\texttt{theodore.papadopoulo@inria.fr} \\
}

\nipsfinalcopy 

\begin{document}

\maketitle

\input{sections/abstract}
\input{sections/introduction}

\input{sections/dl}
\input{sections/jadl}

\input{sections/experiments}
\input{sections/conclusion}

\subsubsection*{Acknowledgments}

This work was supported by the doctoral grant of the region 
Provence-Alpes-C\^{o}te d'Azur and the ANR grants CoAdapt (09-EMER-002-01)
and MultiModel (2010 BLAN 0309 04).

\bibliographystyle{authordate1}

\input{iclr2013_JADL.bbl}
\newpage

\end{document}

%% file: sections/abstract.tex
\begin{abstract}
Dictionary Learning has proven to be a powerful tool for 
many image processing tasks, where atoms
are typically defined on small image patches. As a drawback, 
the dictionary only encodes basic structures.
In addition, this approach treats patches of different locations
in one single set, which means a loss of information when features
are well-aligned across signals.
This is the case, for instance, in 
multi-trial magneto- or electroencephalography (M/EEG). 
Learning the dictionary on the entire signals could make use of 
the alignment and reveal higher-level features. 
In this case, however, small misalignments or phase
variations of features would not be compensated for. In this paper, we propose an 
extension to the common dictionary learning framework to overcome these 
limitations by allowing atoms to adapt their position across 
signals. The method is validated on simulated and real neuroelectric 
data. 
\end{abstract}

%% file: sections/introduction.tex
\section{Introduction}

The analysis of electromagnetic signals induced by brain activity requires sophisticated tools 
capable of efficiently treating redundant, multivariate datasets. Redundancy
originates for example from the spatial dimension as in multi-channel
magneto- or electroencephalography (M/EEG). It can also result from 
repetitions of the same phenomenon, for instance when a stimulus is presented
multiple times to a subject (typically between 50 and 500 times) and the 
neuronal response is recorded; we will refer to this case as multi-trial 
analysis.

In the case of multi-channel data, principal component analysis (PCA)~\cite{pearson1901liii} and 
independent component analysis (ICA)~\cite{comon1994independent}
have been successfully used to decompose the data into a few
waveforms, providing
insight into the underlying neuronal activity and
allowing to enhance the often poor 
signal-to-noise-ratio~\cite{lagerlund1997spatial,makeig1996independent}. They
both use the fact that the data of all channels are recorded 
synchronically such that features appear well-aligned and phase-locked.

This condition typically does not hold for multi-trial analysis though. 
In~\cite{woody1967characterization} a method is 
provided to compensate for temporal jitter across signals, 
but it assumes a single underlying waveform. 
Matching pursuit (MP) algorithms~\cite{mallat1993matching,durka1995analysis}
in turn allow to extract several different features and
have recently been adapted to deal with multi-channel~\cite{durka2005multichannel,gribonval2003piecewise} 
as well as multi-trial~\cite{benar2009consensus} 
M/EEG data by compensating for different
types of variability. However, these methods 
only allow to extract waveforms that have
previously been defined in a dictionary.


In the field of image processing, learning dictionaries directly from the data 
has shown to yield state-of-the-art results 
in several applications~\cite{elad2006image,mairal2008supervised}. Typically 
these dictionaries are learned on small patches and represent the basic structures
of the images, e.g.~edges. When using this technique for neuroelectric multi-trial
analysis though, the framework should be carefully adapted to the properties of the
data:
(i) the waveforms of interest occur approximately at the same time 
across trials; (ii) however, they may have slightly
different time delays (of a small fraction of the signal lenth) or phase variations; 
(iii) the noise-level is
high, partially due to neuronal background activity which is non-Gaussian; (iv) 
data is often limited to a few hundred signals. The latter two properties
make the analysis a difficult problem; it is therefore necessary
to incorporate all prior information about the data, in particular 
(i) and (ii). We note that similar
properties can be found in many other signal processing applications, such
as in other bioelectric or biomagnetic data (e.g.~ECG, EMG). 

We suggest that atoms 
should be learned on the entirety of the signals to provide global high-level
features. The common dictionary learning
formulation as a matrix factorization problem, however, cannot compensate for 
the time delays (ii). Existing extensions known 
as convolutional or shift-invariant sparse coding 
(SISC)~\cite{smith2005efficient,blumensath2006sparse,grosse2007shift,ekanadham2011sparse} learn atoms
that are typically smaller than the signal and can occur at arbitrary and 
possibly multiple positions per signal. This 
framework is very general for our purpose and does not make use of property (i),
the approximate alignment of waveforms. In addition, the SISC framework 
leads to a highly complex algorithm since all shifts of all atoms
are considered in the optimization. The sparse coding step is therefore often handled by 
heuristic preselection of the active atoms, as described in~\cite{blumensath2006sparse}.
But the update of the dictionary elements is also a difficult task, as it involves
solving a convolutional problem~\cite{grosse2007shift}.


In this paper, we present a novel dictionary learning framework that is 
designed specifically to compensate for small temporal jitter of 
waveforms across trials, leading to the name jitter-adaptive dictionary learning (JADL). 
In contrast to SISC, atoms learned by JADL are defined on the entire signal domain and are allowed to shift only
up to a small fraction of the signal length. The most important difference, however,
is a constraint for atoms to occur at most once (i.e.~in one position) per signal, see section \ref{adl:sc}.
On the one hand, this constraint is reasonable since we do not want to encode a waveform with
multiple slightly shifted copies of one atom. On the other hand, it significantly reduces complexity compared to SISC.

An important difference to previous dictionary learning frameworks is the size of the dictionary; while
for image processing dictionaries are often overcomplete, JADL aims at learning only 
a small number of atoms. This is not only desired for easy interpretability, but also because of the  
difficulties introduced by 
(iii) and (iv), that make it infeasible to learn a large number of atoms. 
The ``unrolled'' version of the dictionary, i.e. the set of all allowed shifts of all
atoms may still be large; it therefore makes sense to speak of sparse solutions
with respect to this unrolled dictionary. However, JADL enforces sparsity to a major part 
by the explicit constraint mentioned; sparse
regularization only plays a minor role.

We begin by briefly stating the common dictionary learning
problem after which we will present the theory and implementation details
of JADL. Finally, JADL is evaluated on synthetic and experimental data.

%% file: sections/dl.tex
\section{Dictionary learning: prior art}\label{sec:dict}

A dictionary consists of a matrix
$\D \in \mathbb{R}^{N\times K}$ that contains for columns
the $N$-dimensional column vectors $\{\d_i\}_{i=1}^K$, its \textit{atoms}. 
For a set of signals $\{\x_j \in \R^N\}_{j=1}^M$ 
the problem of finding a sparse \textit{code} $\a_j \in \mathbb{R}^K$
for each $\x_j$ can be formulated as the following minimization problem:
\begin{equation}\label{eq:lasso}
\a_j = \argmin_{\a_j \in \mathbb{R}^K} \frac{1}{2}\left\| \x_j-\D\a_j \right\|_2^2 + \lambda\|\a_j\|_1 \enspace,
\end{equation}
where $\|\cdot\|_1$ denotes the $l_1$-norm and $\lambda>0$ is a regularization parameter.
This problem is known as Lasso \cite{tibshirani1996regression} and can be 
solved efficiently with algorithms such as least angle regression (LARS) \cite{EFRON}
or the proximal methods ISTA \cite{combettes05prox} and its accelerated 
version FISTA \cite{beck2009fast}. 

The case where
$\D$ is not known beforehand but shall be estimated given the signals $\{\x_j\}$,
leads to the dictionary learning problem.
It is a minimization problem over both the dictionary and the 
sparse code, which reads:
\begin{equation}\label{eq:dl}
\begin{split}
&\min_{\D, \a_j} 
\frac{1}{2} \sum_{j=1}^M \left(\left\| \x_j - \D\a_j \right\|_{2}^2 + \lambda \left\| \a_j \right\|_{1}\right),\\
&\text{s.t.} \quad \|\d_i\|_2 = 1, \quad i=1,\ldots,K,
\end{split}
\end{equation}
where the latter constraint prevents atoms from growing arbitrarily large. 

Most algorithms tackle this non-convex problem iteratively by 
alternating between the convex subproblems: (i) the sparse coding (with $\D$ fixed) and (ii) the dictionary
update ($\{\a_j\}_{j=1}^M$ fixed). The first such algorithm was provided in the pioneer work on 
dictionary learning in \cite{Olshausen1997}. Many
alternatives have been proposed, such as
the method of optimal directions (MOD) in \cite{engan1999method},
$K$-SVD \cite{Aharon2006}, or more recently an
online version \cite{Mairal2010} to handle large datasets.

%% file: sections/jadl.tex
\section{Jitter-adaptive dictionary learning}\label{sec:jadl}

This section introduces the main contribution of this paper: a novel technique 
designed to overcome the limitations of purely linear signal decomposition
methods such as PCA, ICA, and dictionary learning. 

We suppose that atoms present in a signal
can suffer from unknown time delays, which we will refer to as \textit{jitter}.
This type of variability addresses the issue of varying latencies of 
transient features as well as oscillations with different phases
across trials. 

This issue is very important for interpretation of M/EEG data, to answer 
fundamental questions such as the link between evoked responses and 
oscillatory activity~\cite{hanslmayer-klimesch-etal:07,mazaheri-jensen:08}, 
the correlation between single-trial activity and behavioral data~\cite{jung-makeig-etal:01}, 
the variability of evoked potentials such as the P300~\cite{holm-ranta-aho-etal:06}. 
Cross-trial variability is also a precious source of information for simultaneous 
EEG-fMRI interpretation~\cite{benar-schon-etal:07}.

We therefore provide a dictionary learning framework in which atoms 
may adapt their position across trials. This framework can
be generalized to address other kinds of variability; the
shift operator introduced below can simply be replaced by the
desired operators. The entire framework remains the same, only the 
formula for the update of the dictionary needs to be adapted
as described in Section~\ref{sec:dict_up}.

\subsection{Model and problem statement}
Our model is based on the hypothesis that the set of signals of interest $\{\x_j\}_{j=1}^M$
can be generated by a dictionary $\D=\{\d_i\}_{i=1}^K$ with few atoms $K$ in the following way: Given
a set of shift operations $\Delta$
\footnote{There are different possibilities to define these shifts.
As our entire framework is valid even for arbitrary linear transforms, we do not specify the choice
at this point. While circular shifts, i.e.~$\delta^{n}(\d)=d^n$ for $n \in \N$ and 
$\d^n[i]:=\d[(i-n)\text{mod} N]$, result in a slightly simpler formulation
of the dictionary update and may have minor computational advantages, 
they can introduce unwanted boundary effects. Our implementation actually uses atoms defined on a slightly larger 
domain ($N+S-1$ sample points) than the signals, this way avoiding circular shifts. For the sake of simplicity, 
however, we here assume atoms to be defined on the same domain as the signals. Although the 
right way to handle boundary effects can be an important question, 
it is out of the scope of this paper to
discuss this issue in detail. In our experiments, we found the impact of 
the concrete definition of the $\delta$ to be small. }
, for every $j$ there exist coefficients
$a_{ij} \in \R$ and shift operators $\delta_{ij} \in \Delta$, such that 
\begin{equation}\label{eq:model}
\x_j = \sum_{i=1}^K a_{ij}\delta_{ij}(\d_i) \enspace .
\end{equation}
We assume that $\Delta$ contains only small shifts relative to 
the size of the time window. Now, we can formulate
the jitter-adaptive dictionary learning problem 
\begin{equation}\label{eq:adl_dl}
\begin{split}
&\min_{\d_i,a_{ij}, \delta_{ij}}  \sum_{j=1}^M \left( \frac{1}{2}\left\|\x_j 
- \sum_{i=1}^K a_{ij}\delta_{ij}(\d_i)\right\|_2^2 + \lambda \left\| \a_j\right\|_1 \right),\\
&\text{s.t.} \quad \left\| \d_i\right\|_2 = 1, \quad \delta_{ij} \in \Delta, \quad i=1,\ldots,K, \quad j=1,\ldots,M.
\end{split}
\end{equation} 
Note that for $\Delta=\{\Id\}$ this reduces to Eq.~(\ref{eq:dl}), 
the problem is thus also non-convex and we solve it similarly using alternate minimizations.
%
%
%

%
\subsection{Implementation}\label{sec:imp}
The algorithm that we propose for solving Eq.~(\ref{eq:adl_dl}) is based on
an implementation in~\cite{Mairal2010} for common dictionary learning, which 
iteratively alternates between (i) sparse coding and (ii) dictionary update. We
adapt the algorithm least angle regression (LARS) used for solving (i) 
to find not only the coefficients $\{a_{ij}\}$ but also the latencies 
$\{\delta_{ij}\}$. For (ii), block coordinate descent is used.
The entire procedure is summarized in Algorithm 1 (notation 
and details are explained in the following).

\begin{algorithm}
  \caption{Jitter-Adaptive Dictionary Learning}
  \label{alg:adl}
  \begin{algorithmic}[1]
    \REQUIRE $\text{signals } \{\x_1,\x_2,\ldots,\x_M\}, \text{shift operators } \Delta, K \in \mathbb{N}, \lambda \in \mathbb{R}, $
    \STATE Initialize $\D=\{\d_1,\d_2,\ldots,\d_K\}$ 
    \REPEAT
      \STATE \text{Set up ``unrolled'' dictionary $\D^S$ from $\D$}
      \STATE \textit{\textbf{Sparse coding} (solve using modified LARS)}      
      \FOR{$j=1$ to $M$}
	\STATE 
	\begin{equation*}
	\a^{S}_j \leftarrow \argmin \frac{1}{2} \left\| \x_j- \D^S\a^S_j \right\|_2^2 
  +\lambda \left\| \a^S_j \right\|_1, \quad \text{s.t.} \quad \left\|\a_j^{S,i}\right\|_0 \leq 1
	\end{equation*}
      \ENDFOR
      \STATE \text{Convert $\{\a^{S}_j\}$ to $\{a_{ij}\}, \{\delta_{ij}\}$}
      \STATE \textit{\textbf{Dictionary update} (solve using block coordinate descent) }
      \STATE
      \begin{equation*}
	\D \leftarrow \argmin_{\{\d_{i}\}_{i=1}^K}  \sum_{j=1}^M \frac{1}{2}\left\|\x_j 
- \sum_{i=1}^K a_{ij}\delta_{ij}(\d_i)\right\|_2^2, \quad \text{s.t.} \quad \|\d_i\|=1
      \end{equation*}
    \UNTIL{convergence}
  \end{algorithmic}
\end{algorithm}

\subsubsection{Sparse coding}\label{adl:sc}


When $\D$ is fixed, the minimization Eq.~(\ref{eq:adl_dl}) can be solved independently for each 
signal $\x_j$, 
\begin{equation}\label{eq:adl_sc}
\min_{a_{ij}, \delta_{ij}} \frac{1}{2}\left\|\x_j 
- \sum_{i=1}^K a_{ij}\delta_{ij}(\d_i)\right\|_2^2 + \lambda \left\| \a_j\right\|_1.\\
\end{equation}
We now rewrite this problem into a form similar to the Lasso, which allows us
to solve it using a modification of LARS. Let us first define an ``unrolled'' version
of the dictionary containing all possible shifts of all its atoms; this is 
given by $\D^S = \{\delta(\d): \d \in \D, \delta \in \Delta\}$,
a matrix of dimension ${N\times KS}$, where $S=|\Delta|$ is the number of 
allowed shifts. The decomposition in Eq.~(\ref{eq:adl_sc}) can now be
rewritten as a linear combination over the unrolled dictionary 
\begin{equation*}
\sum_{i=1}^K a_{ij}\delta_{ij}(\d_i) = \D^S\a^S_{j},
\end{equation*}
%
where $\a^S_{j} \in \R^{KS}$ denotes the 
corresponding coefficient vector. This vector is extremely sparse; in fact, 
each subvector $\a^{S,i}_{j}$ of $\a^S_{j}$ that contains the coefficients
corresponding to the shifts of atom $\d_i$ shall
maximally have one non-zero entry. If 
such a non-zero entry exists, its position indicates which 
shift was used for atom $\d_i$.
Now Eq.~(\ref{eq:adl_sc}) can be rewritten as
\begin{eqnarray}
\a^{S}_j \leftarrow & \argmin&\frac{1}{2} \left\| \x_j- \D^S\a^S_j \right\|_2^2 +\lambda \left\| \a^S_j \right\|_1,\label{eq:asc_mod}\\ 
& \text{s.t.}& \quad \left\|\a_j^{S,i}\right\|_0 \leq 1, \quad i=1,\ldots,K. \label{eq:asc_con}
\end{eqnarray}
%
Clearly, Eq.~(\ref{eq:asc_mod}) is the Lasso, but
the constraint (\ref{eq:asc_con}) leads to a non-convex problem. Therefore
the modification of the LARS that we propose below only
guarantees convergence to a local minimum. 

The LARS algorithm~\cite{EFRON} 
follows a stepwise procedure; in each step the coefficient of one atom is selected 
to change from ``inactive'' to ``active'' (i.e.~it changes from zero to non-zero)
or vice versa. In order to enforce the constraint (\ref{eq:asc_con}), 
we make the following modification. When a coefficient is selected for activation, 
we determine the index $i$ such that this coefficient
lies in $\a_j^{S,i}$. We then block all other coefficients contained in the subvector
$\a_j^{S,i}$ such that they cannot get activated in a later step. In the 
same manner, 
we unblock all entries of $\a_j^{S,i}$ when its active coefficient is deactivated.

As mentioned in the introduction, the constraint (\ref{eq:asc_con}), which is the 
main difference to SISC, helps to reduce
the complexity of the optimization. In fact, each time an atom is activated, all 
its translates are blocked and do not need to be considered in the following steps, 
which facilitates the calculation. In addition, maximally $K$ steps have to be 
performed (given that no atom is deactivated in a later step, which we observed to occur rarely).  
As suggested for example in~\cite{grosse2007shift} the inital correlations of the shifted atoms with the signal can 
be computed using fast convolution via FFT, which speeds up computation in the case of a large
number of shifts $S$.

\subsubsection{Dictionary update}\label{sec:dict_up}
For the dictionary update, its unrolled version
cannot be used as this would result in updating different shifts of the 
same atom in different ways. Instead, the shifts have to be explicitly included
in the update process. 
We use block coordinate descent to iteratively solve the constrained 
minimization problem
\begin{equation*}
  \d_k = \argmin_{\d_k} \sum_{j=1}^M \frac{1}{2}\left\|\x_j 
- \sum_{i=1}^K a_{ij}\delta_{ij}(\d_i)\right\|_2^2, 
\quad \text{s.t.} \quad \|\d_k\|_2 = 1
\end{equation*}
for each atom $\d_k$. This can be solved in two steps, the solution
of the unconstrained problem by differentiation followed by normalization. 
This is summarized by
\begin{eqnarray}
\widetilde{\d_k} &=& \sum_{j=1}^M a_{kj}\delta_{kj}^{-1} \left(
\x_j - \sum_{i\neq k}a_{ij}\delta_{ij}(\d_i)
\right), \label{eq:update} \\
\d_k &=& \frac{\widetilde{\d_k}}{\|\widetilde{\d_k}\|_2}. \nonumber 
\end{eqnarray}
with $\delta^{-1}$ the opposite shift of $\delta$. 
As in~\cite{Mairal2010}, we found that one update loop through all of the 
atoms was enough to ensure fast convergence of Algorithm \ref{alg:adl}.

The only difference of this update compared to common dictionary learning are
the shift operators $\delta_{ij}$. In contrast to the sparse coding step, the
jitter adaptivity therefore does not increase the complexity for this step. When
high efficiency of the algorithm is needed, this fact can be used by employing
mini-batch or online techniques~\cite{Mairal2010}, which increase the frequency
of dictionary update steps with respect to sparse coding steps.


We note that in Eq.~(\ref{eq:update}) we assumed the shifts to be circular;
being orthogonal transforms, i.e. $\delta \delta^t = \Id$, they provide
for a simple update formula. In the case of non-circular shifts or 
other linear operators, the inverse $\delta_{kj}^{-1}$ in 
the update Eq.~(\ref{eq:update}) needs to be replaced 
by the adjoint $\delta_{kj}^{t}$. In addition, the rescaling function
$\psi = (\sum_{j=1}^M a_{kj}^2 \delta_{kj}\delta^t_{kj})^{-1}$ has to 
be applied to the update term.

\subsubsection{Hyperparameters and initial dictionary}

As mentioned before, the optimal number $K$ of atoms
will typically be lower than for dictionary learning in 
image processing frameworks; this
is due to the high redundancy of the electromagnetic brain signals as well as 
the adaptive property of the atoms. When oscillatory activity is sought, 
$\Delta$ should contain shifts of 
up to the largest period expected to be found.  
The choice of $\lambda$ plays a less important role than in common dictionary 
learning; most of the sparsity is induced directly by the constraint (\ref{eq:asc_con}).

The choice of the initial dictionary can be crucial for the outcome,
due to the non-convex nature of the problem. Especially when the
initial dictionary already produces a small sparse
coding error, the algorithm may converge to a local minimum that
is very close to the initialization. While this property 
allows us to incorporate a priori knowledge by initializing
with a predefined dictionary, this also provides the risk of preventing the
algorithm from learning interesting features if they are ``far'' from the initial dictionary.

\begin{figure}
\centering
\parbox{0.45\linewidth}{%
\includegraphics[width=\linewidth, height=6cm]{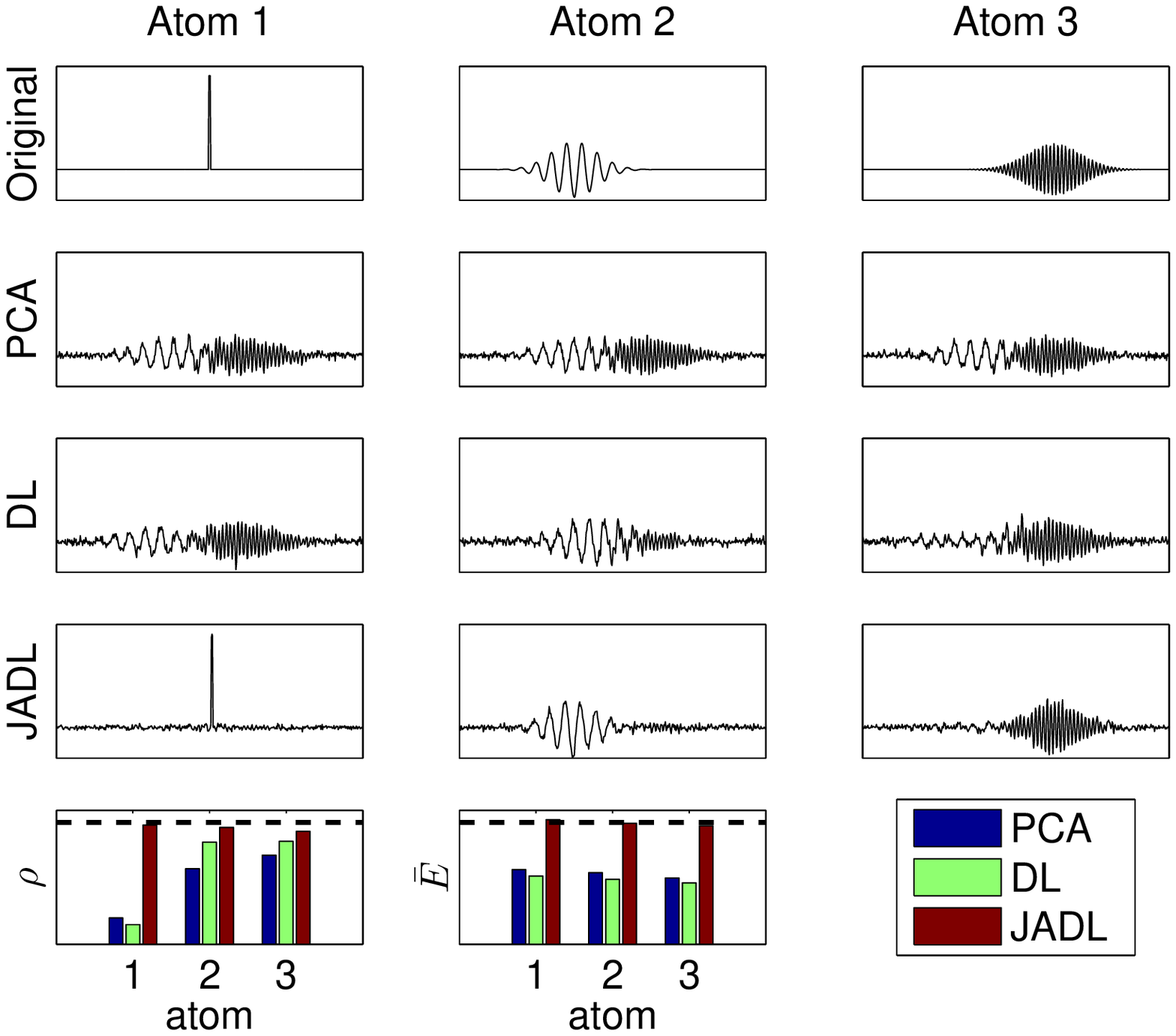}
\caption{Original dictionary and reconstructions
with PCA, DL, and JADL, respectively; row 5: similarity $\rho$ and 
average energy across signals $\bar{E}$ for each atom, the dashed line
marks the value 1 in both plots. }\label{fig:dict_sim}
}
\qquad
\begin{minipage}{0.45\linewidth}%
\includegraphics[width=\linewidth, height=6cm]{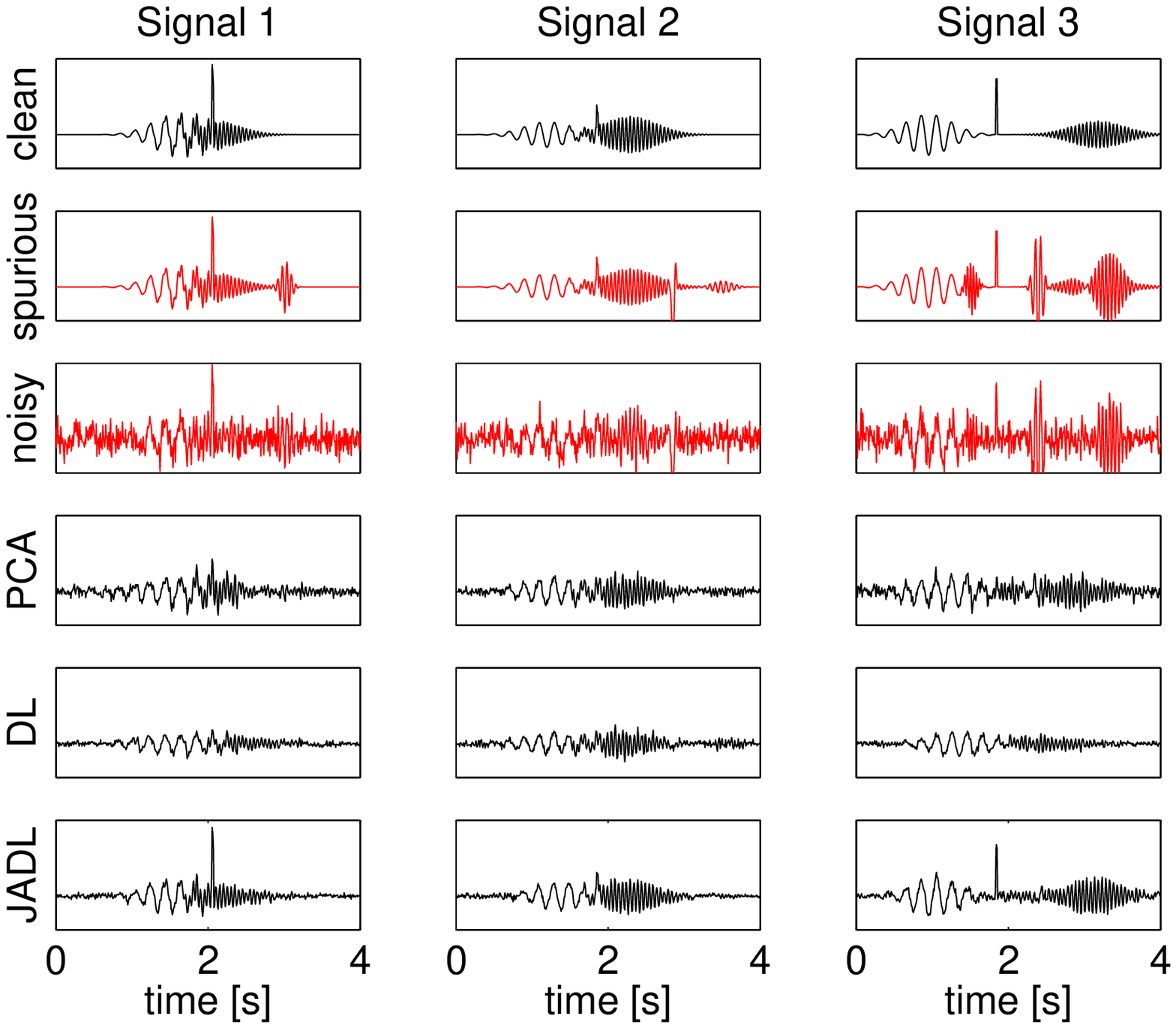}
\caption{Clean, noisy and denoised signals, row 1: clean signals; row 2: signals plus spurious events;
row 3: signals above plus white noise, row 4-6:
denoised signals using PCA, DL, and JADL, respectively. }\label{fig:signals_rec}
\end{minipage}%
\end{figure}

When no dictionary is given a priori, initializations often found in the literature
include random values as in~\cite{Olshausen1997}
as well as random examples~\cite{Aharon2006} or linear combinations (e.g. PCA)
of the input signals. In order to keep
the bias towards the initialization as small as possible, we favor random 
initializations independent from the signals.

%% file: sections/experiments.tex
\section{Experiments}\label{sec:exp}

Jitter-adaptive dictionary learning (JADL) is next evaluated 
on synthetic and on experimental data.
Its performance is compared to results obtained with 
principal component analysis (PCA)\cite{pearson1901liii}
and dictionary learning (DL);
for the latter we used the open-source software package 
SPAMS\footnote{\texttt{http://spams-devel.gforge.inria.fr/}}
whose implementation
is described in \cite{Mairal2010}.

\subsection{Synthetic data}\label{sec:exp_sim}

\begin{floatingtable}[r]
{
\input{images/table_similar.tex}
}
\vspace{-2mm}
\caption{First three rows: average similarity $\bar{\rho}$ of original and reconstructed atoms;
last two rows: parameter $\lambda$ used for reconstruction.}\label{tab:similarity}
\end{floatingtable}

\textbf{Generating the dictionary:} 
First, a dictionary $\D$ with $K=3$ normalized atoms was defined, as shown in
Figure \ref{fig:dict_sim} (first row). The first atom is a delta-shaped spike
and the other two are oscillatory features. The length
of the time window was chosen as $4$ seconds and the sampling rate as 
128 Hz, hence $N=512$.

\textbf{Generating the signals:} From the dictionary, $M=200$ signals were generated according to the model
Eq.~(\ref{eq:model}). The coefficients $\a_{ij}$ and shifts $\delta_{ij}$ 
were drawn independently from Gaussian distributions with mean $\mu=1$
and standard deviation $\sigma=0.3$ for the coefficients and $\mu=0, \sigma = 0.2$ s
for the shifts. Three examples are shown in 
the first row of Figure \ref{fig:signals_rec}. These signals were then
corrupted by two types of noise: (i) to every signal between $0$ and 
$3$ oscillatory events were added, their amplitudes, frequencies,
temporal supports and positions in time were drawn randomly
(Fig.~\ref{fig:signals_rec}, row 2);
(ii) then white Gaussian noise with 
a resulting average SNR (energy of clean signals/energy of noise) of 0.790
was added (row 3).

\textbf{Reconstructions:}
Performance of the three methods PCA, DL, and JADL given the noisy signals
(Fig.~\ref{fig:signals_rec}, row 3) was measured on both
their ability of recovering the original dictionary and the denoising
of the signals. We 
performed reconstruction for different dictionary sizes $K$. For DL and JADL
we chose the $\lambda$ that gave the best results; we noticed relatively small
sensitivity of the methods to the choice of $\lambda$,
especially for small values of $K$.

\textbf{Recovering the dictionary:}
For PCA, the dictionary atoms
were defined as the first $K$ principal components. 
For JADL, the set $\Delta$ was defined to contain all time shifts 
of maximally $\pm 0.6$ seconds, resulting in $S=128 \text{ Hz} \cdot 1.2 \text{ s}\approx 154$ allowed
shifts.
For each reconstructed dictionary, the three atoms that
showed the highest similarity $\rho$ to the original atoms 
were used to calculate the average similarity $\bar{\rho}$. 
$\rho$ was defined as the maximal correlation of the reconstructed atom
and all shifts of maximally $\pm 0.6$ seconds of the original atom.
The values $\bar{\rho}$ are shown for PCA, DL, and
JADL in Table \ref{tab:similarity} for each $K$ (for $K>12$ 
we observed decreasing similarity values for all three methods). 
The similarity for JADL is 
significantly higher than for PCA and DL; its optimal value is 
obtained for $K=3$ atoms, but it remains high for larger $K$, showing
its robustness to overestimation of original atoms. Note that dictionaries 
obtained by PCA for different values of $K$ always have their first
atoms in common, hence the constant similarity values.

\begin{floatingtable}[r]
{
\begin{small}\begin{tabular}{|l|c|c|c|c|c|c|c|c|c|}
\hline
&\textbf{K=1}&\textbf{K=2}&\textbf{K=3}&\textbf{K=4}&\textbf{K=5}&\textbf{K=6}&\textbf{K=8}&\textbf{K=10}&\textbf{K=12}\\\hline
\textbf{$\epsilon$ PCA}&0.871&0.750&0.638&0.539&0.522&0.508&\textbf{0.502}&0.537&0.570\\\hline
\textbf{$\epsilon$ DL}&0.869&0.747&0.635&0.535&0.515&0.498&\textbf{0.487}&0.505&0.539\\\hline
\textbf{$\epsilon$ JADL}&0.505&0.283&\textbf{0.214}&0.230&0.277&0.284&0.317&0.325&0.330\\\hline
\textbf{$\lambda$ DL}&0.05&0.05&0.05&0.05&0.05&0.05&0.1&0.1&0.1\\\hline
\textbf{$\lambda$ JADL}&0.05&0.2&0.1&0.2&0.2&0.3&0.4&0.5&0.5\\\hline
\end{tabular}
\end{small}}
\vspace{-2mm}
\caption{Relative $l_2$-error $\epsilon$ produced by each method for different $K$;
the last two rows show the values of $\lambda$ used for DL and JADL.}\label{tab:errors}
\end{floatingtable}

For each method, the $K$ giving the highest similarity was determined
and the three atoms with largest $\rho$-values of the corresponding 
dictionaries are shown in Figure \ref{fig:dict_sim} (row $2$ - $4$).
The atoms found by PCA and DL contain only mixtures of the oscillatory atoms, the 
spike does not appear at all. JADL succeeds in separating the three atoms; their
shapes and average energy across signals $\bar{E}$ are very close to the originals, 
as shown in the bar plots. 

\textbf{Denoising the signals:}
For PCA, denoising was performed by setting the coefficients
of all but the first $K$ components to zero. 
For DL and JADL, the noisy signals were encoded over 
the learned dictionaries
according to Eq. (\ref{eq:lasso}) and (\ref{eq:adl_sc}), respectively.
Table
\ref{tab:errors} shows the average relative $l_2$-errors $\epsilon$ of
all denoised signals with respect to the original ones.
For each method, three denoised signals for optimal $K$
are shown in the last three rows of Figure \ref{fig:signals_rec}.
Despite the larger dictionary size ($K=8$) in the case of PCA and DL, 
the spike could not be reconstructed due to its jitter across signals.
In addition, the locations of the oscillatory events in the signals
denoised with JADL correspond better to their true locations 
than it is the case for PCA and DL, especially
in the last signal shown. Finally, the
white noise appears more reduced for JADL, which is due to its smaller dictionary size.
All three methods succeeded in removing the spurious events from the signals.
\subsection{Real data}\label{sec:exp_real}


Local field potentials (LFP) were recorded with an intra-cranial electrode
in a  Wistar-Han rat, in an animal model of epilepsy.   
Bicuculline (a blocker of inhibition) 
was injected in the cortex in order to elicit epileptic-like discharges. 
Discharges were selected visually on the LFP traces, and the data was 
segmented in 169 epochs centered around the spikes. To simplify the analysis,
the epochs were scaled with a constant factor such that the maximal
$l_2$-energy among epochs was equal to $1$.

\begin{figure}
\centering
\subfigure{
\includegraphics[width=0.18\linewidth, height=2cm]{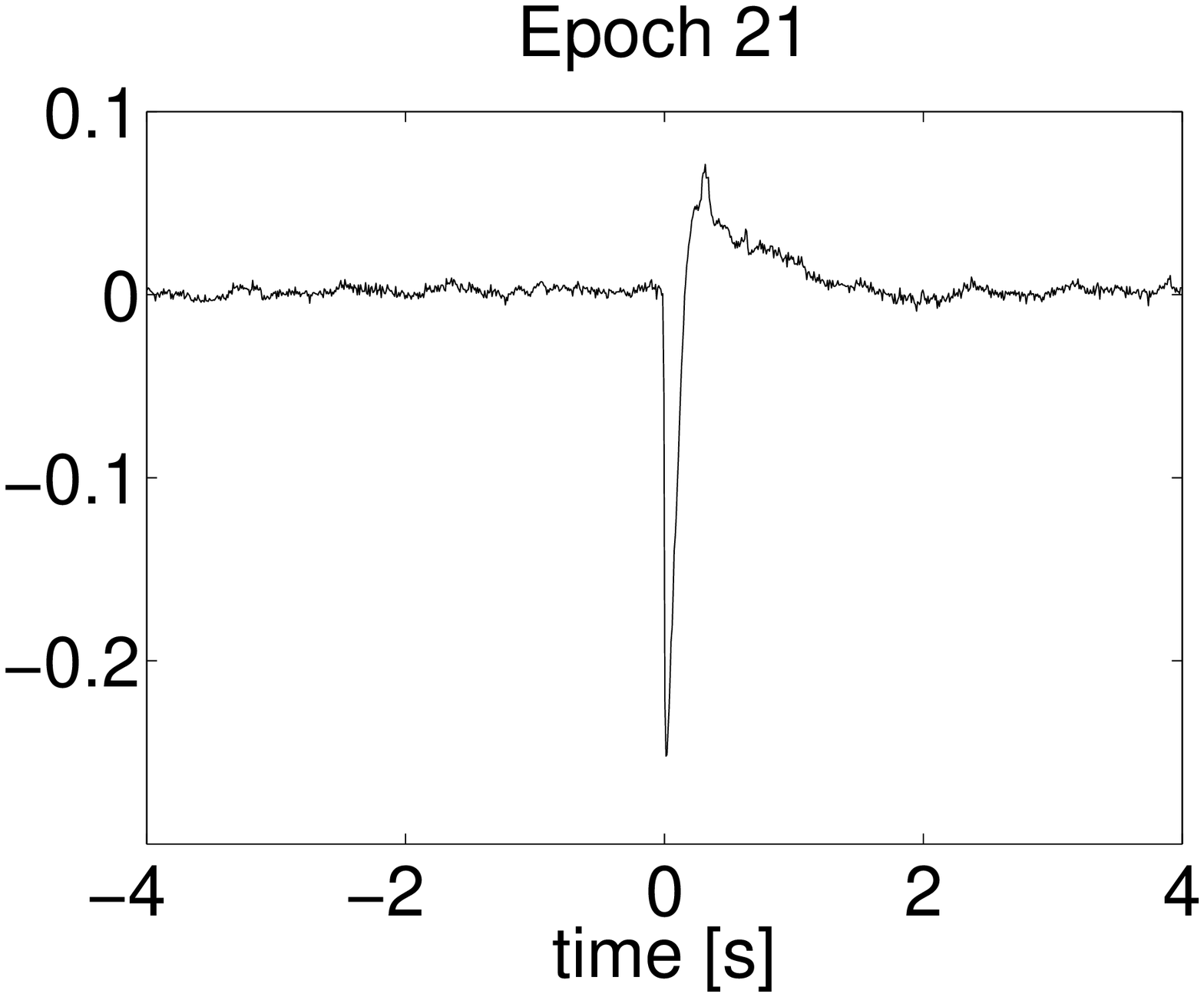}
}
\subfigure{
\includegraphics[width=0.18\linewidth, height=2cm]{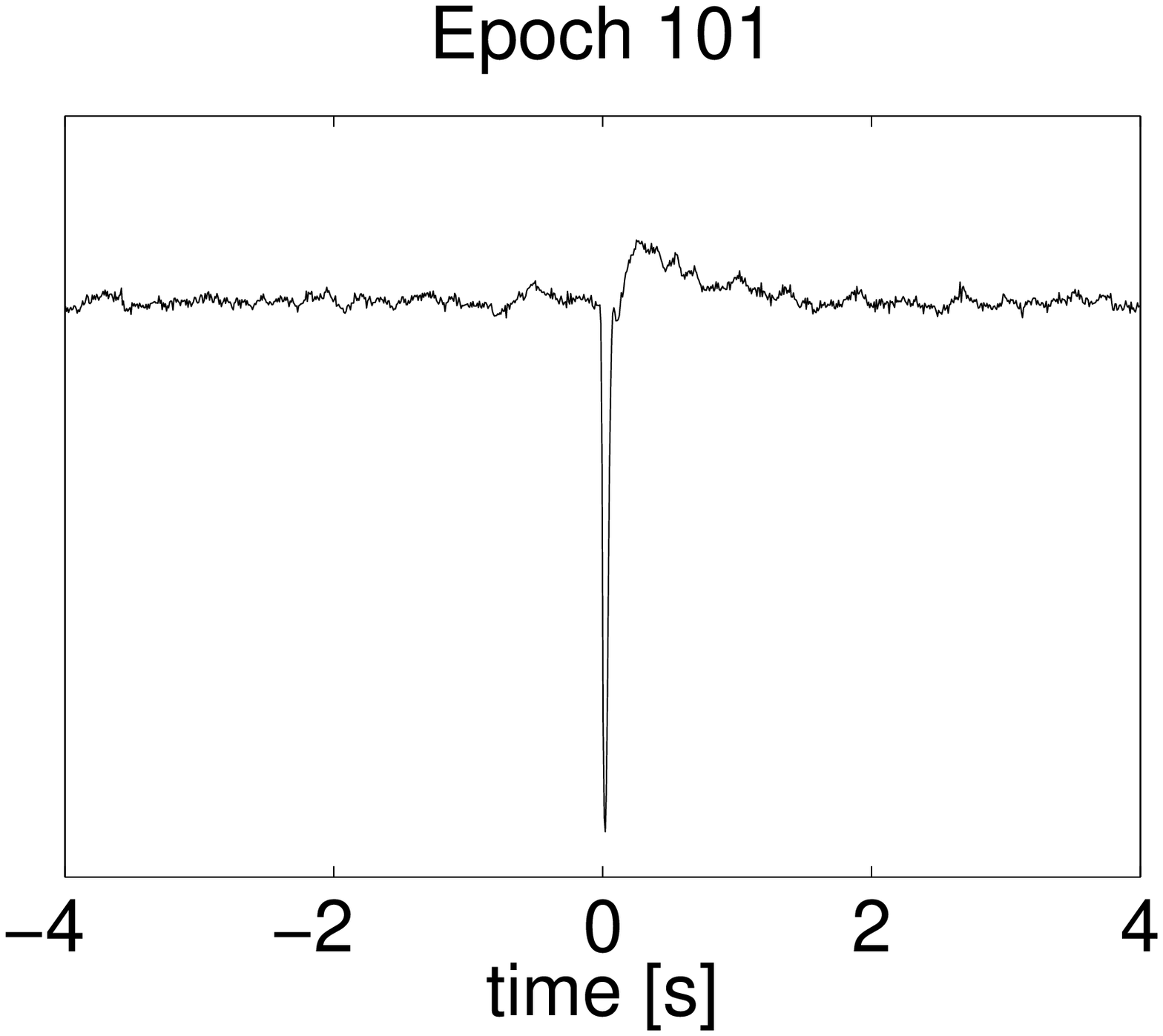}
}
\subfigure{
\includegraphics[width=0.18\linewidth, height=2cm]{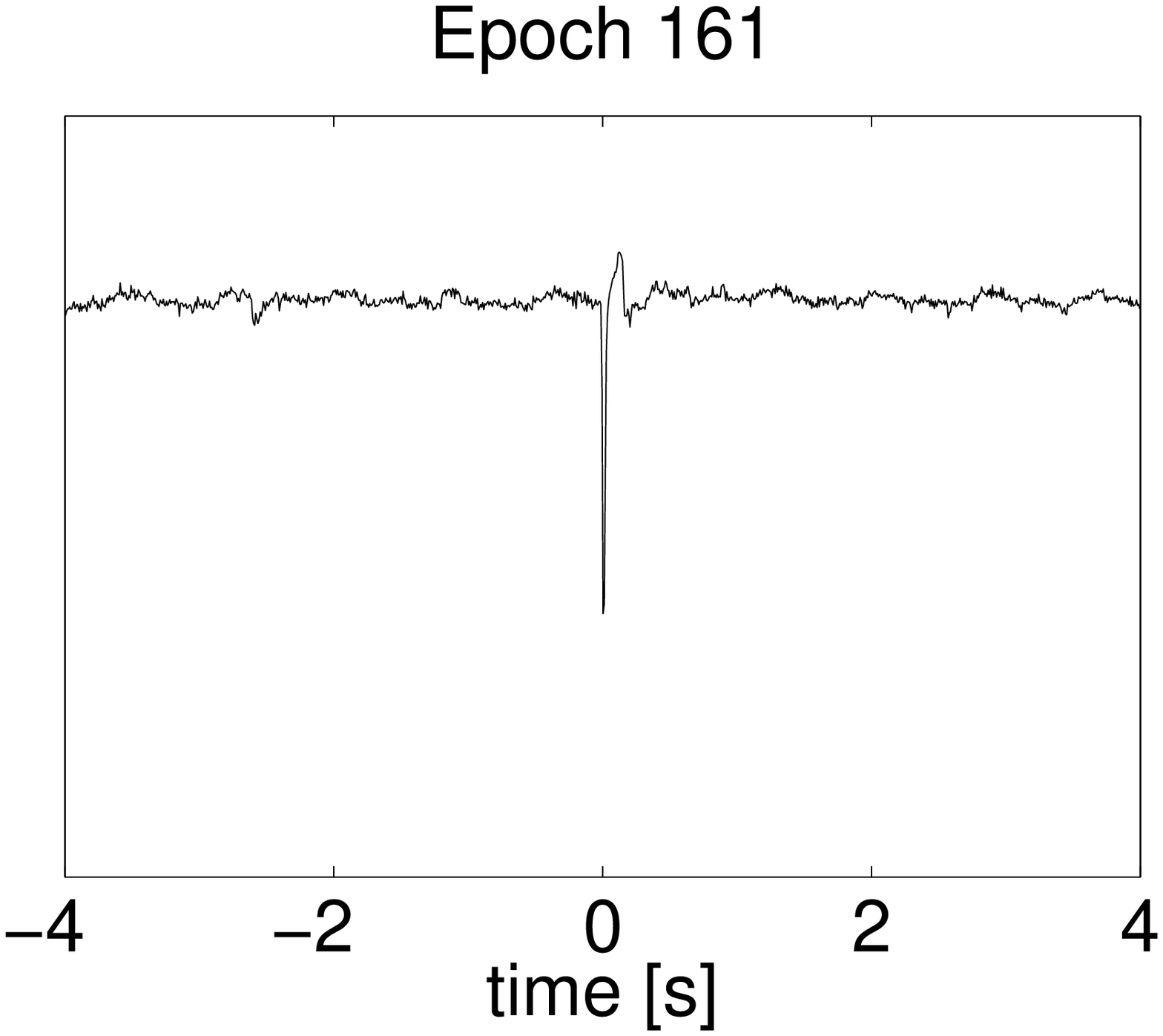}
}
\subfigure{
\includegraphics[width=0.18\linewidth, height=2cm]{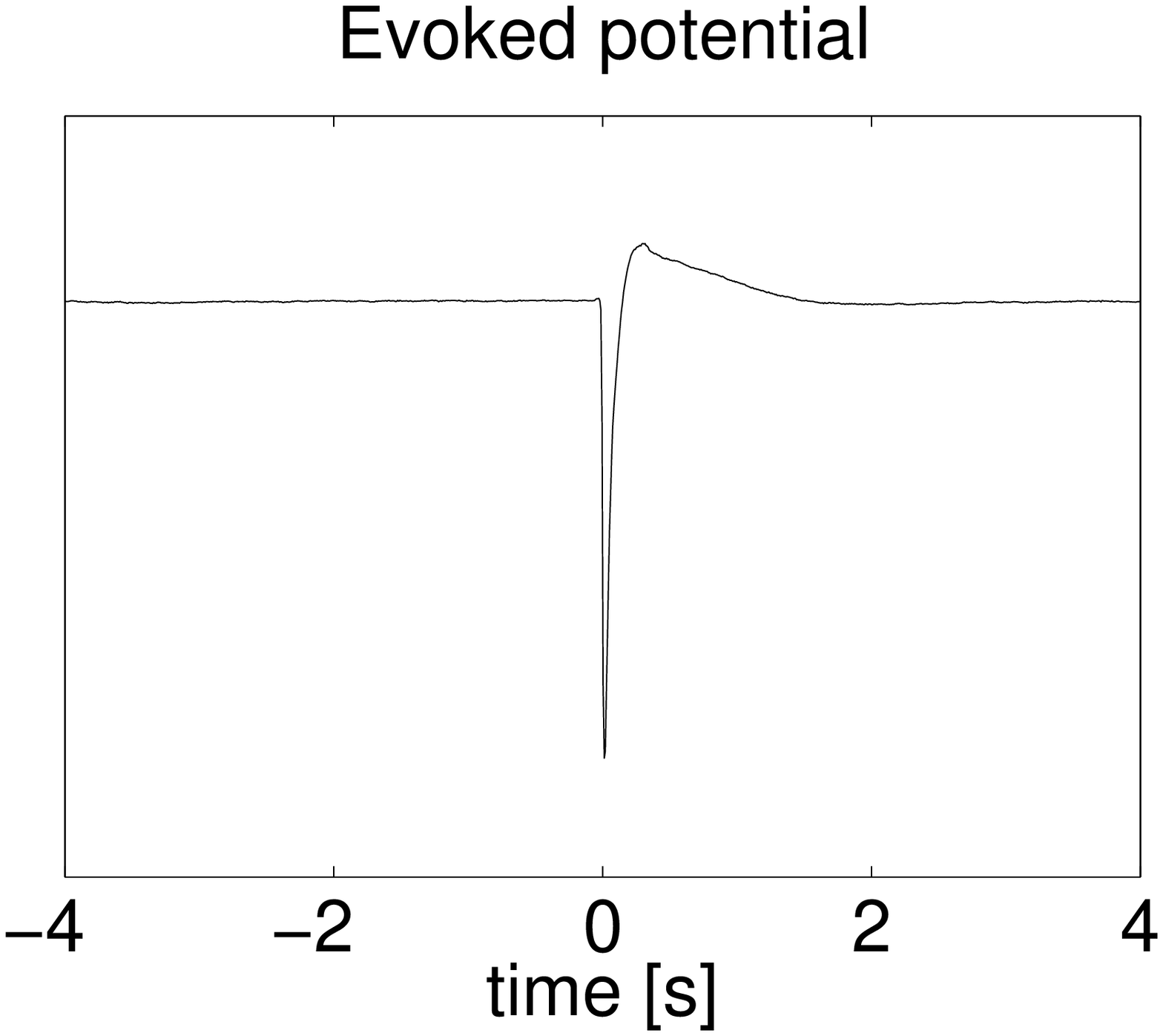}
}
\caption{Different epochs of the local field potential (LFP) showing spikes with 
decreasing energy; the evoked potential (last) is the average over all 
epochs.}\label{fig:samp_spikes}

\end{figure}

Three examples of epochs are shown in Figure \ref{fig:samp_spikes}, as well as the
evoked potential, measured as the average over the epochs. The only structure visible
to the eye is a spike with changing shape and decreasing energy across epochs.

\begin{figure}
\centering
\includegraphics[width=0.85\linewidth, height=7cm]{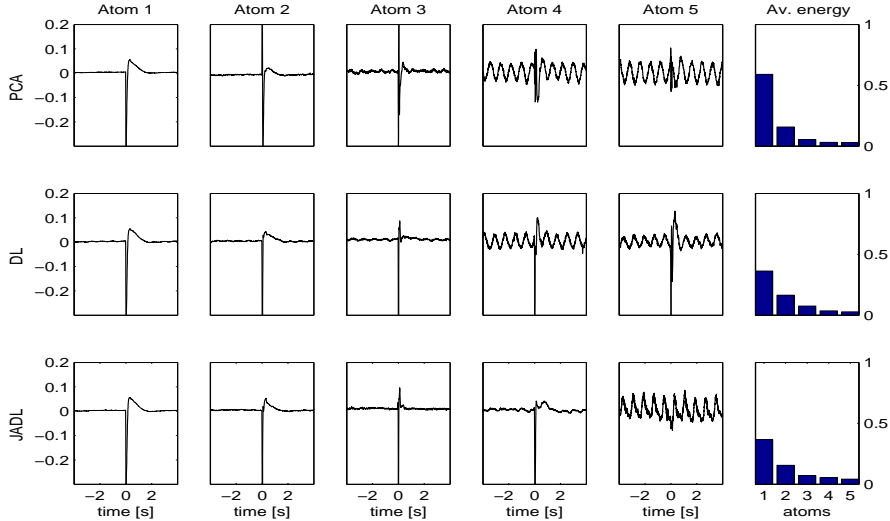}
\caption{Dictionaries learned on LFP epochs using PCA, DL, and JADL.}\label{fig:res_comp}
\end{figure}

\textbf{Learning the dictionary:}
Five normalized atoms were learned on the data with PCA, DL 
and JADL; see Figure \ref{fig:res_comp}. They were ordered
in descending order of their average energy across epochs.

All three methods produce for their first atom a spike that resembles the 
evoked potential (Figure~\ref{fig:samp_spikes}); also, all
methods reveal an oscillatory artifact around 1.2 Hz which is not visible
in the evoked potential. However, while the oscillations in the PCA and DL
dictionaries
are encoded in two atoms (4 and 5) that differ mostly in phase, they are concentrated in 
atom 5 for JADL. Additionally, in the case of JADL the oscillations are almost
completely separated from the spike, only a small remainder is still visible. 
This shows the need of PCA and DL for several atoms to compensate for phase shifts 
while JADL is able to associate oscillations with different phases; moreover JADL 
makes use of the varying phases to clearly separate transient from oscillatory events. 
In addition, we can observe a smoothing effect in the case of PCA and DL: 
the oscillations look very much like sinusoids whereas
atom 5 in JADL shows a less smooth, rather spiky structure.

\begin{figure}
\centering
\subfigure[Absolute coefficients of atoms learned with JADL.\label{subfig:res_coeffs}]{
  \includegraphics[width=0.3\linewidth, height=4cm]{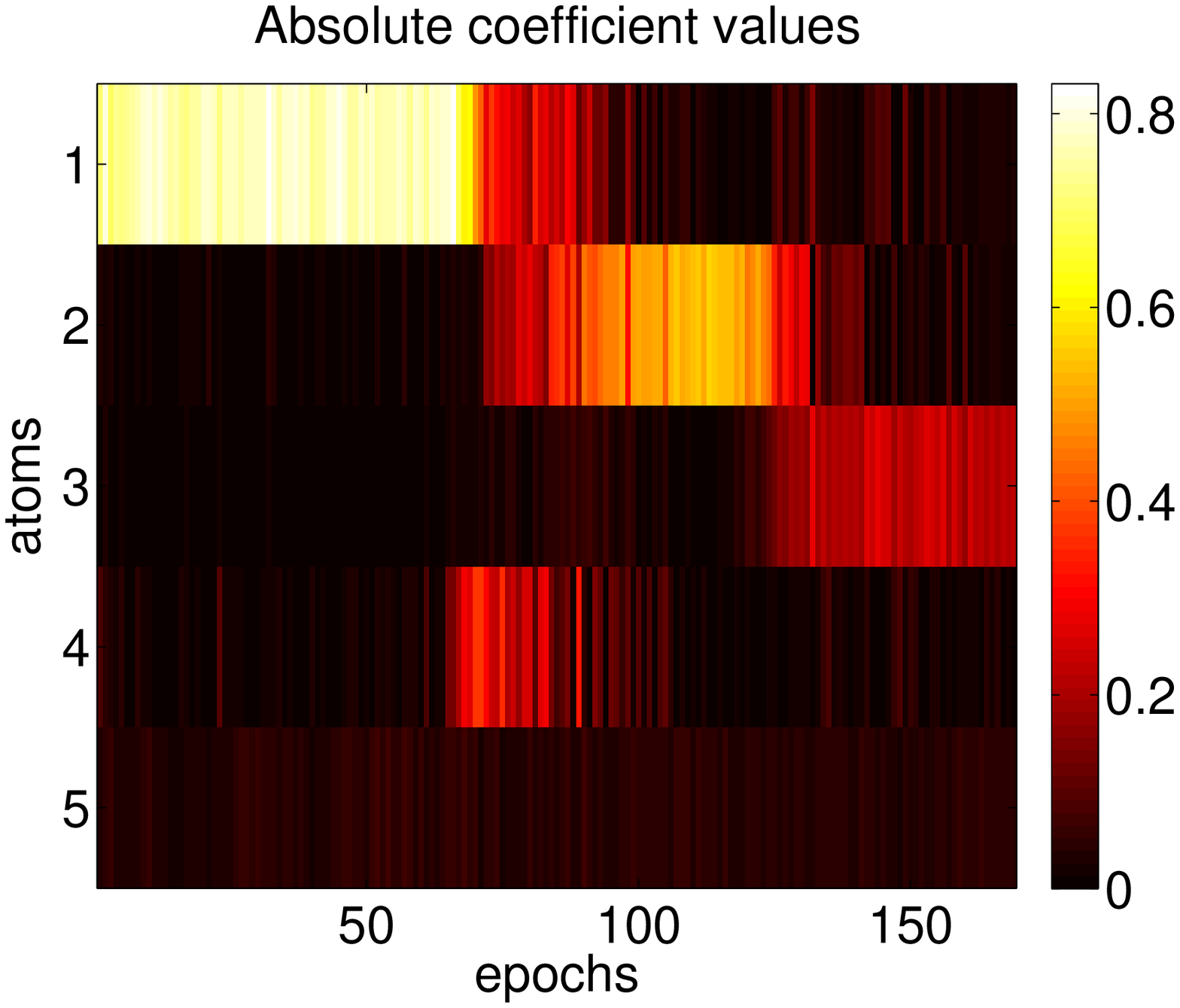}  
}
\quad
\subfigure[Distribution of latencies for each atom.\label{subfig:res_lats}]{
  \includegraphics[width=0.3\linewidth, height=4cm]{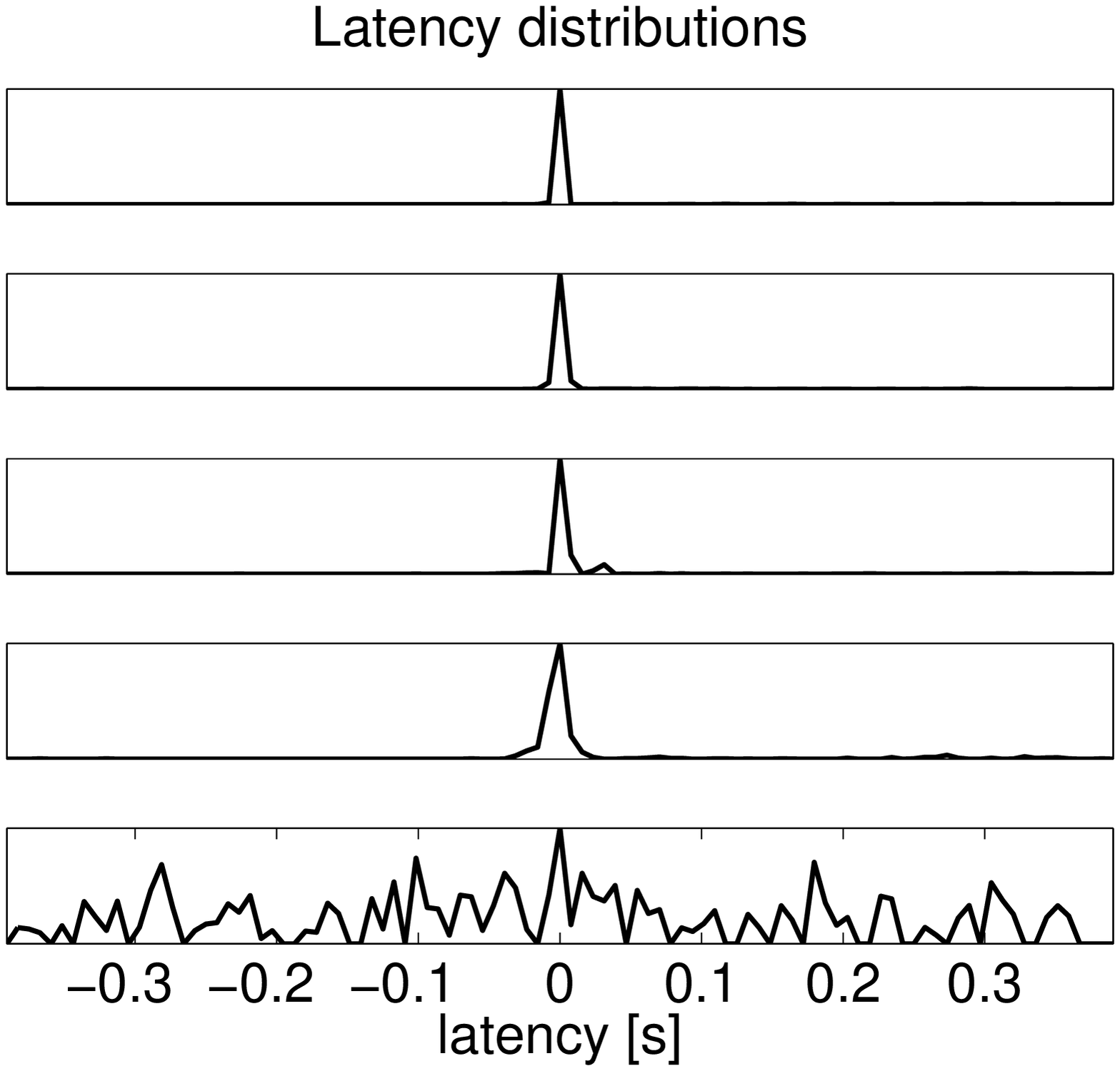}
}
\quad
\subfigure[Computation times for different values of $S$ and $K$.\label{subfig:times}]{
  \includegraphics[width=0.3\linewidth, height=4cm]{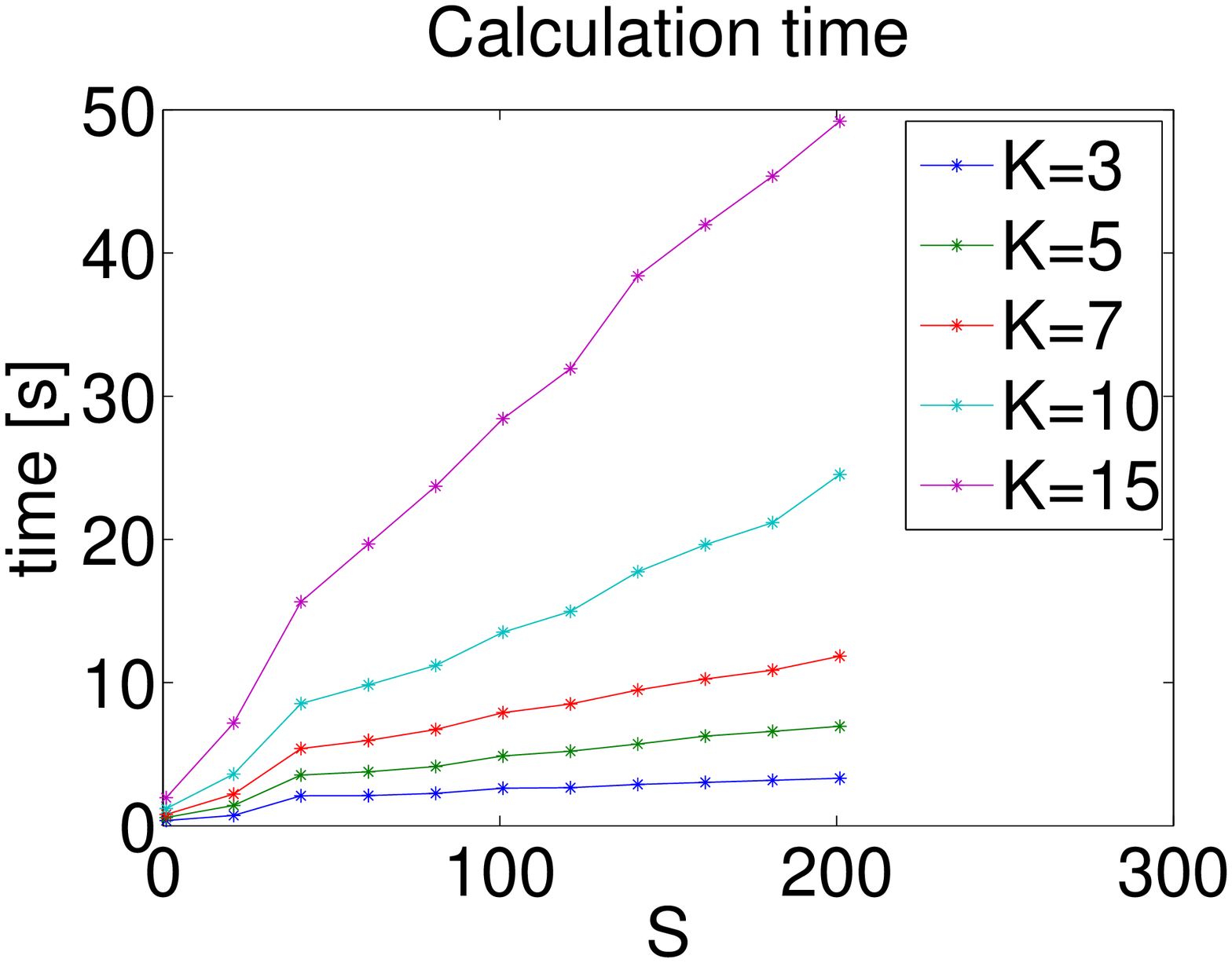}
}
\caption{Code visualization and computation times.}\label{fig:code_vis}
\end{figure}

\textbf{Interpreting the code:}
We visualized the coefficients and the shifts obtained by 
decomposing all the epochs over the dictionary learned with JADL, 
see Figure \ref{fig:code_vis}.
Interestingly, each of the first three spikes in the dictionary 
is active during a contiguous set of epochs during which the other
atoms are hardly used. This allows to segment the epochs into
three sets of different spike shapes. The fourth atom is only
active during a few epochs where the dominant atom is changing
from the first to the second spike. The oscillatory artifact (atom 5) 
in contrast shows very low but constant activity across all epochs. 

The latency distributions (Fig.~\ref{subfig:res_lats}) can provide
further insight into the data. The highly peaked distribution for the first atoms
gives evidence of the accurate alignment of the spikes. The last atom
shows shifts in the whole range from $-0.4$ to $0.4$ seconds,
indicating that the phases of the oscillations were uniformly distributed across
epochs. 

\textbf{Computation times:}
The dictionary and the code above were obtained for 189 iterations of JADL
after which the algorithm converged. The time of computation on a laptop (Intel Core CPU, 2.4 GHz) 
was 4.3 seconds. As JADL
is designed for datasets of similar complexity to the one investigated here,
computation time should not be an issue for offline analysis. However, applications
such as M/EEG based brain computer interfaces (BCI) may require computations
in real time. We remark that our implementation has not yet been optimized and could
be speeded up significantly by parallelization or online techniques 
mentioned in section \ref{sec:dict_up}. If training data 
is available beforehand, the dictionary may also be calculated offline
in advance; the sparse encoding of new data over this dictionary 
then only takes up a small fraction of the training time and can be performed online. 

Figure \ref{subfig:times} illustrates the effect of changing the number of 
atoms $K$ and shifts $S$ on computation time $t$; for each calculation 200 iterations were
performed. We can see that $t$ is linearly correlated with $S$ but increases
over-linearly with $K$: while both, $S$ and $K$ 
affect the size of the unrolled dictionary $\D^S$, an increase of $S$ is 
handled more efficiently by using calculation advantages described in section \ref{adl:sc};
e.g., the non-smooth behavior of the curves at $S=40$ results 
from the fact that for $S>40$ an FFT-based convolution is used. 
In addition, the dictionary update step only uses the compact dictionary 
$\D$ whose size does not increase with $S$. If the additional 
factor in computation time due to $S$ is still not acceptable, 
$\Delta$ may be subsampled by introducing a minimal
distance between shifts; the tradeoff is a less exact description
of the atoms latencies. 

%% file: images/table_similar.tex
\begin{small}\begin{tabular}{|l|c|c|c|c|c|c|c|}
\hline
&\textbf{K=3}&\textbf{K=4}&\textbf{K=5}&\textbf{K=6}&\textbf{K=8}&\textbf{K=10}&\textbf{K=12}\\\hline
\textbf{$\bar{\rho}$ PCA}&0.522&0.522&0.522&0.522&0.522&0.522&0.522\\\hline
\textbf{$\bar{\rho}$ DL}&0.563&0.566&0.598&\textbf{0.615}&0.589&0.595&0.581\\\hline
\textbf{$\bar{\rho}$ JADL}&\textbf{0.955}&0.954&0.946&0.911&0.931&0.881&0.801\\\hline
\textbf{$\lambda$ DL}&0.001&0.005&0.001&0.001&0.005&0.050&0.2\\\hline
\textbf{$\lambda$ JADL}&0.001&0.005&0.005&0.01&0.005&0.1&0.4\\\hline
\end{tabular}
\end{small}

%% file: sections/conclusion.tex
\section{Conclusion}
In this paper, a novel method for the analysis and processing of
multi-trial neuroelectric signals was introduced. The method
was derived by extending the dictionary learning framework, 
allowing atoms to adapt to jitter across trials; hence the name
jitter-adaptive dictionary learning (JADL). It was shown 
how the resulting non-convex minimization problem can be
tackled by modifying existing algorithms used in common
dictionary learning.

The method was validated on synthetic and experimental data, both 
containing variability in latencies of transient
and in the phases of oscillatory events. The results obtained
showed to be superior to those of common dictionary learning 
and PCA, both in recovering the underlying dictionary and 
in denoising the signals. The evaluation furthermore demonstrated 
the usefulness of JADL as a data exploration tool, 
capable of extracting global, high-level features of the signals
and giving insight into their distributions.